# MaskBot: Real-time Robotic Projection Mapping with Head Motion Tracking

Miguel Altamirano Cabrera
Skolkovo Institute of Science and Technology (Skoltech)
Moscow, Russia
miguel.altamirano@skoltech.ru

Igor Usachev
Skolkovo Institute of Science and Technology (Skoltech)
Moscow, Russia
igor.usachev@skoltech.ru

Juan Heredia
Skolkovo Institute of Science and Technology (Skoltech)
Moscow, Russia
juan.heredia@skoltech.ru

Jonathan Tirado
Skolkovo Institute of Science and Technology (Skoltech)
Moscow, Russia
jonathan.tirado@skoltech.ru

Aleksey Fedoseev
Skolkovo Institute of Science and Technology (Skoltech)
Moscow, Russia
aleksey.fedoseev@skoltech.ru

Dzmitry Tsetserukou
Skolkovo Institute of Science and Technology (Skoltech)
Moscow, Russia
d.tsetserukou@skoltech.ru

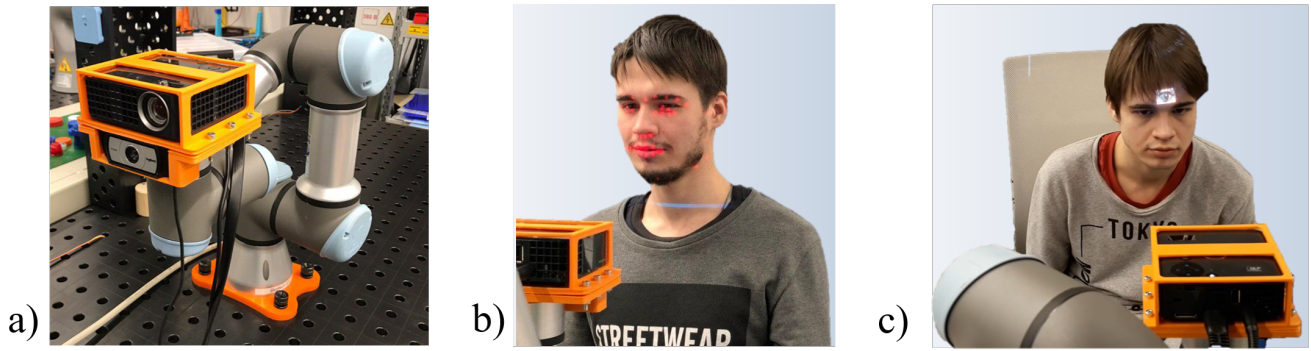

Figure 1: MaskBot system: a) The module with the projector and the webcam located at the 6DoF UR3 robot's end-effector. b) Face detection and projection of 68 points. c) The projection mapping on the user's face.

## ABSTRACT
The projection mapping systems on the human face is limited by the latency and the movement of the users. The area of the projection is restricted by the position of the projectors and the cameras. We are introducing MaskBot, a real-time projection mapping system operated by a 6 Degrees of Freedom (DoF) collaborative robot. The collaborative robot locates the projector and camera in normal position to the face of the user to increase the projection area and to reduce the latency of the system. A webcam is used to detect the face and to sense the robot-user distance to modify the projection size and orientation. MaskBot projects different images on the face of the user, such as face modifications, make-up, and logos. In contrast to the existing methods, the presented system is the first that introduces a robotic projection mapping. One of the prospective applications is to acquire a dataset of adversarial images to challenge face detection DNN systems, such as Face ID.

## 1 INTRODUCTION
The projection mapping has been used in many fields to solve some important problems of object visualization in augmented reality. Inami et al. [1] introduced projection on objects covered by retroreflective material to allow the users to handle objects of arbitrary shape. The users look at bright images projected on the surface of the object covered with retroreflective material. This approach recreates the shape of the object that is observed and creates the illusion to see a different one. This method can be applied to generate many illusions and to change the visual shape of objects.

However, the face mapping projection to create a different patterns on the human face does not have to apply a material that covers the face.

The technology introduced by Matrosov et al. generates a projection on the floor as a screen to interact with the drone by feet gestures [2]. The novelty of this work is in the intuitive interface based on RGB-D camera and projector integrated in a flying robot. The proposed technology allows to recognize the gestures of the user's foot and generate corresponding response of the application. The LightAir is not suitable for the dataset collection as it is challenging for hovering drone with stabilized position. CollMot Entertainment [3] developed a system that uses drones to generate an enormous screen in the sky, and then laser projectors draw on that screen.

The latency required in the face projection mapping is much higher than the required for non-deformable objects. Bermano et al. [5] proposed a system to considerably reduce the latency in the face projection mapping. In this project, the facial expressions are accurately detected and can be predicted with Kalman filters. However, the equipment used by them is too sophisticated, and the position of the face is not possible to be changed. Bokaris et al. introduced a system that used more straightforward approach. This system makes a real-time facial projection mapping to create the effect of make-up on the face of the users [4]. However, projection mapping is limited to the area of the projector.

We are introducing MaskBot, a novel system that increases the projection avoiding any distortion caused by the user head misalignment. The technology behind this project is a robot arm with 6 DoF and CV algorithm, which controls the orientation of the projector and the camera regardless of the user's head posture. The position of the human head is tracked with a compact camera. When human moves the head, the robotic projector will always trace the posture to emit light towards the normal surface. The safety of the human-robot interaction is secured by the collaborative robot sensitivity to the contact.

## 2 MASKBOT CONFIGURATION

MaskBot provides a real-time projection mapping on moving face and on other objects. The main components of our system are a 6 DoF collaborative robot UR3 from Universal Robots, a Logitech Webcam C930e, and a ultra-compact mobile projector Dell M115HD. The webcam and the projector are attached to the robot's end-effector as indicated in the Fig. 1.

The image captured by the camera is processed with a Python script. Face detection algorithm provides face center coordinates. These coordinates are recalculated into the distance between the face and the camera. Next, the robot moves towards human face. In parallel, the projection algorithm lays an image on the particular area of the face. In this way MaskBot provides an extra work space due to the robot movements. In addition, the system is able to change the projection plane and produce projection on other objects.

## 3 FACE PROJECTION MAPPING

We are using face landmark detection by the model in *dlib* trained by Sagonas et al. [6] where 68 landmarks are detected on the user's face. These point are used to locate different objects on the face and create deformation to fit on the human face.

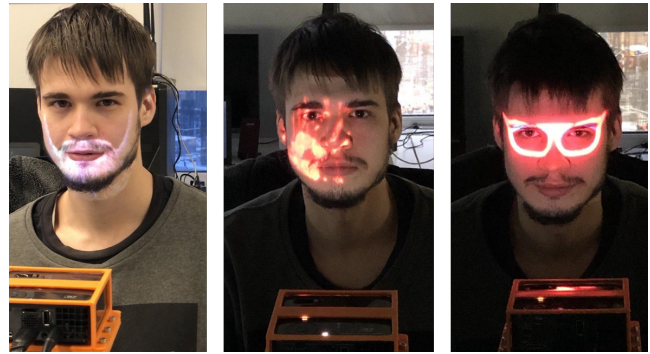

**Figure 2: Projection of the beard, hand, glasses on the face.**

The picture projected on the face of the users is created using the *OpenCV* library according to the calculated angle for the orientation. The distance from the user's face to the MaskBot system is calculated by the relation between the face wide detected by CV and the real face wide.

During demonstration, the users stand in front of the system. In a first intense, a camera-projector calibration is going to be carried. In the central part of the demonstration, a series of masks will be projected on the face, and each one is going to be changed by gesture recognition from the user. The second robot will move the tablet showing the face of the user with the projected image. The visitors will be able to draw their own patterns on the user's face.

In the future work, a camera will capture the image of the human face modified by the projected image and the dataset of adversarial images will be collected. We can also envision a novel communication system where human communicates with no need to aline the face towards cameras as it is the case with Skype. Also, the technology can be applied in Industry 4.0 when robots communicate the messages through projectors, e.g., they can illuminate QR-code, instructions, and content of the parcel to the partner robot and workers.